\documentclass[10pt,twocolumn,letterpaper]{article}

\usepackage[pagenumbers]{wacv} 

\usepackage[accsupp]{axessibility}
\usepackage{adjustbox}
\usepackage{arydshln}
\newcommand{\modelname}{MiniGPT-5\xspace}

\definecolor{wacvblue}{rgb}{0.21,0.49,0.74}
\usepackage[pagebackref,breaklinks,colorlinks,allcolors=wacvblue]{hyperref}

\def\wacvPaperID{70} 
\def\confName{WACV}
\def\confYear{2026}

\title{Interleaved Vision-and-Language Generation via Generative Voken}

\author{
    Kaizhi Zheng$^{1}$\thanks{Equal contribution} \quad
    Xuehai He$^{1}$\footnotemark[1] \quad
    Xin Eric Wang$^{1,2}$ \\
    $^{1}$University of California, Santa Cruz \quad
    $^{2}$University of California, Santa Barbara \\
    {\tt\small \{kzheng31@ucsc.edu, xhe89@ucsc.edu, ericxwang@ucsb.edu\}}
}

\begin{document}
\maketitle
\begin{abstract}
  The effectiveness of Multimodal Large Language Models (MLLMs) demonstrates a profound capability in multimodal understanding.
However, the simultaneous generation of images with coherent texts is still underdeveloped. 
Addressing this, we introduce a novel interleaved vision-and-language generation method, centered around the concept of ``generative vokens". These vokens serve as pivotal elements contributing to coherent image-text outputs. Our method is marked by a unique two-stage training strategy for description-free multimodal generation, which does not necessitate extensive descriptions of images. We integrate classifier-free guidance to enhance the alignment of generated images and texts, ensuring more seamless and contextually relevant multimodal interactions.
 Our model, \modelname, exhibits substantial improvement over the baseline models on multimodal generation datasets, including MMDialog and VIST. The human evaluation shows \modelname is better than the baseline model on more than 56\% cases for multimodal generation, highlighting its efficacy across diverse benchmarks. Project page: \url{https://eric-ai-lab.github.io/minigpt-5.github.io/}.
\end{abstract}

\section{Introduction}
\begin{figure}[!t]
     \centering
     \includegraphics[width=0.5\textwidth]{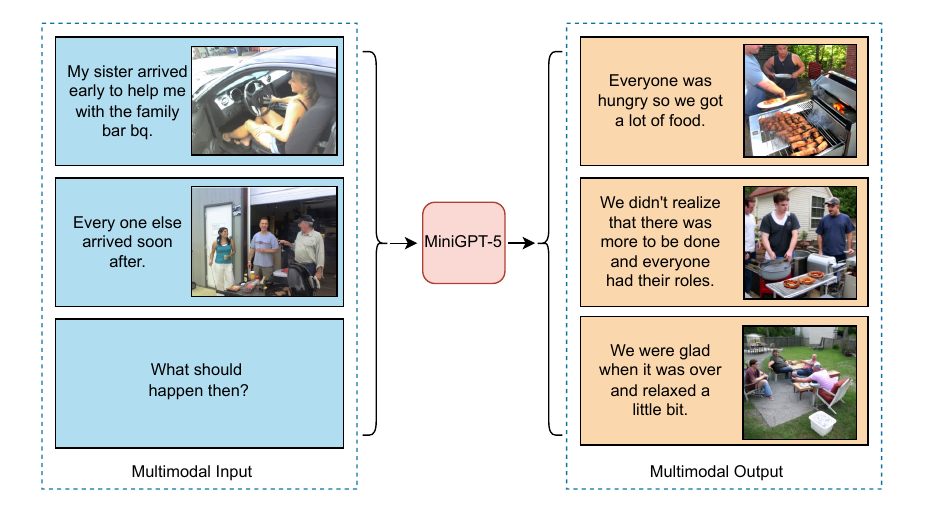}
     \caption[Overview of \modelname]{\modelname is a unified model for interleaved vision-and-language comprehension and generation. Besides the original multimodal comprehension and text generation abilities, \modelname can provide appropriate, coherent multimodal outputs. }
     \label{fig:teaser}

\end{figure}

The development of large-scale vision-and-language models is significantly impacting a wide range of fields like automated dialogue systems and digital content creation. With the surge in research and development in this domain, the current state-of-the-art Large Language Models (LLMs)~\cite{openai2023gpt4, vicuna2023, training_lm} and vision-and-language models such as~\cite{visualchatgpt,li2023blip,frozen,alayrac2022flamingo} fall short in generating coherent multimodal outputs. This limitation becomes particularly evident in tasks that demand an integrated handling of vision and language, essential for the next generation Large Language Models (LLMs). 

Our work, as illustrated in Fig.~\ref{fig:teaser}, seeks to address these shortcomings by enhancing the integration of text and image generation in LLMs. The challenges in developing a multimodal LLM capable of interleaved vision and language generation are manifold. First, LLMs typically lack mechanisms to directly produce images, prompting us to introduce “generative vokens” that bridge the gap between textual and visual feature spaces. Second, the constraint of data scarcity, especially in vision-and-language tasks~\cite{sharma2018conceptual} lacking extensive detailed captions of images~\cite{huang2016visual}, like descriptions of the elements inside images, is countered by our unique description-free training approach. Third, maintaining both image-text and image-image consistency poses a significant challenge, which we address through dual-loss strategies. Finally, as we push forward the boundaries with LLMs, the large memory requirements urge us to devise more efficient end-to-end strategies and create an efficient training pipeline accessible for the community, especially in downstream tasks. 

Specifically, to overcome these challenges, we present \modelname, a novel approach for interleaved vision-and-language generation. By combining Stable Diffusion with LLMs through special visual tokens~\cite{tan2020vokenization} -- ``generative vokens", we develop a new approach for multimodal generation. Our two-stage training methodology emphasizes a description-free foundational phase, enabling effective model training even with limited caption-grounded images. This strategy, distinct from existing works, pivots on generic stages free from image annotations. To ensure that the generated text and images are in harmony, our dual-loss strategy comes into play, further enhanced by our innovative generative voken approach and classifier-free guidance. Our parameter-efficient fine-tuning strategy optimizes training efficiency and addresses memory constraints.

As shown in Fig.~\ref{fig:overview}, leveraging ViT (Vision Transformer) and Qformer~\cite{li2023blip}, alongside Large Language Models, we adapt multimodal inputs into generative vokens, seamlessly combined with the high-resolution Stable Diffusion 2.1 model~\cite{rombach2021highresolution} for context-aware image generation. Incorporating images as auxiliary input with instruction tuning approaches and pioneering both the text and image generation loss, we amplify the synergy between text and visuals. We experiment on the CC3M~\cite{sharma2018conceptual}, VIST~\cite{huang2016visual}, and MMDialog~\cite{feng2022mmdialog} datasets. Notably, \modelname shows superior performance across the two multimodal generation datasets.

In summary, our contributions are primarily threefold:
\begin{itemize}
\item We introduce a novel framework that leverages “generative vokens” to unify LLMs with Stable Diffusion, facilitating interleaved vision-and-language generation without relying on detailed image descriptions. We bridge the modality gap and improve the generation quality by using the loss of the latent diffusion model, the text generation loss, and the caption alignment loss together during training.
\item We propose a new two-stage training strategy for description-free multimodal generation. The first stage focuses on extracting high-quality text-aligned visual features from large text-image pairs, while the second stage ensures optimal coordination between visual and textual prompts during generation. The inclusion of classifier-free guidance during training enhances the overall generation quality.
\item 
\modelname achieves significant improvements over baseline methods on interleaved vision-and-language datasets, including VIST and MMDialog, and comparable results to the state-of-the-art on the single text-image pair dataset, CC3M. 
The human evaluation further shows that, compared with the two-stage baseline, \modelname can provide better generation in perspectives of appropriate texts (55\%), high-quality images (53\%), and coherent multimodal outputs (56\%).
\end{itemize}

\section{Related Work}

\noindent\textbf{Multimodal Large Language Models}
As Large Language Models (LLMs) become increasingly impactful and accessible, a growing body of research has emerged to extend these pretrained LLMs into the realm of multimodal comprehension tasks~\citep{zhu2023minigpt,li2023blip,instructblip,openai2023gpt4,li2023otter,alayrac2022flamingo,li-etal-2023-lavis,bai2025qwen2}. For example, to reproduce the impressive multimodal comprehension ability in GPT-4~\citep{openai2023gpt4}, MiniGPT-4~\citep{zhu2023minigpt} proposes a projection layer to align pretrained vision component of BLIP-2~\citep{li2023blip} with an advanced open-source large language model, Vicuna~\citep{vicuna2023}. In our work, we utilize the MiniGPT-4 as the base model and extend the model’s capabilities to multimodal generation.

\vspace{1ex}
\noindent\textbf{Text-to-Image Generation~}
To transform textual descriptions into their corresponding visual representations, text-to-image models~\citep{reed2016generative,dhariwal2021diffusion,saharia2022photorealistic,rombach2021highresolution,rombach2022high,gu2023photoswap,nichol2021glide,ramesh2021zero,yu2022scaling,chang2023muse,esser2024scaling} design algorithms to bridge the gap between textual information and visual content. A notable recent contribution is Stable Diffusion V2~\citep{rombach2021highresolution}, which employs a diffusion process to generate conditional image features and subsequently reconstructs images from these features. Our research aims to leverage this pretrained model, enhancing its capabilities to accommodate both multimodal input and output.

\vspace{1ex}

\noindent\textbf{Multimodal Generation with Large Language Models.~}
Recent work augments LLMs to \emph{generate} as well as \emph{understand} across modalities via retrieval-augmented decoders, tokenized image streams, or learned visual-token interfaces~\citep{sun2021multimodal,aiello2023jointly,ge2023planting,koh2023generating,yu2023scaling,wu2023nextgpt,dong2023dreamllm,sun2023emu}. 
CM3Leon~\citep{yu2023scaling} uses a retrieval-augmented, decoder-only architecture for text$\leftrightarrow$image tasks. 
Emu~\citep{sun2023emu} converts images to 1D features with EVA-CLIP~\citep{sun2023eva} and fine-tunes LLaMA~\citep{touvron2023llama} to autoregress over text and \emph{text-encoder–conditioned} image features (requiring subsequent T2I fine-tuning).
NextGPT~\citep{wu2023nextgpt}, GILL~\citep{koh2023generating}, and SEED~\citep{ge2023planting} map “visual tokens” into the \emph{text} feature space of a pretrained diffusion model (via encoder–decoder or Q-Former variants).
A parallel line builds \emph{unified} autoregressive models that couple understanding and synthesis in a single network by introducing large vocabularies of image tokens and performing extensive full-model (re)training, e.g., Chameleon~\citep{team2024chameleon}, Emu3~\citep{wang2024emu3}, and Show-o~\citep{xie2025showo}.
Unlike unified models, \modelname is \emph{modular and parameter-efficient}, reusing pretrained MLLM and diffusion backbones while tuning only a lightweight feature mapper and LoRA. 
Its key novelty is \emph{generative vokens} that project LLM states directly into the diffusion \emph{conditional} space and are trained with a latent diffusion loss; paired with \emph{description-free} interleaved fine-tuning and \emph{CFG over vokens}, this yields efficient, long-horizon interleaved generation with minimal backbone changes.

\section{Method}
\label{sec:method}
\begin{figure*}[!t]
     \centering
     \includegraphics[width=0.9\textwidth]{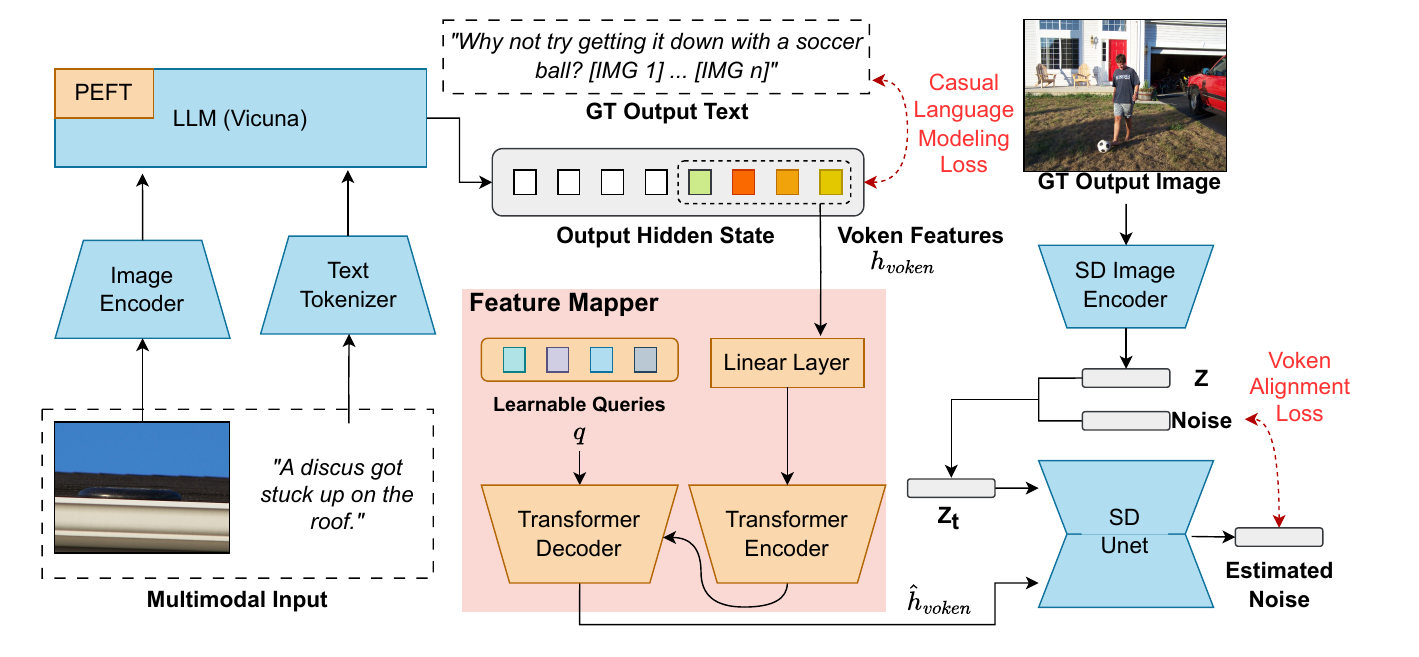}
     \caption{The overview structure of \modelname pipeline. We leverage the pretrained multimodal large language model (MiniGPT-4) and text-to-image generation model (Stable Diffusion 2.1) to create a unified multimodal generation pipeline. The input image encoder includes a ViT, Qformer, and linear layer, pretrained by MiniGPT-4. The orange blocks include learnable parameters, while the blue blocks are fixed during training. More details can be found in Section~\ref{sec:method}.}
     \label{fig:overview}

\end{figure*}

In order to endow Large Language Models with multimodal generation capabilities, we introduce a new framework that integrates pretrained multimodal Large Language Models and text-to-image generation models. Central to our approach is the introduction of “generative vokens”, special visual tokens that effectively bridge the textual and visual domains during the training process. Additionally, we implement a two-stage training method combined with a classifier-free guidance strategy to enhance the quality and coherence of generated outputs. Fig.~\ref{fig:overview} provides an overview of our model structure. \modelname primarily consists of two modules: the Multimodal Understanding module, utilizing the pretrained multimodal large language model (MiniGPT-4) for handling multimodal inputs, and the Multimodal Generation module, employing Stable Diffusion for generating visual outputs.

\subsection{Multimodal Understanding Module}
Recent advancements in multimodal Large Language Models, such as MiniGPT-4~\cite{zhu2023minigpt}, have primarily concentrated on multimodal comprehension, enabling the processing of images as sequential input. 
The Integrated Vision-Language Encoding Module is designed to extend the capabilities of LLMs from mere comprehension to active generation in multimodal contexts. Generative vokens play a crucial role in this module, enabling the translation of raw visual inputs into a format that LLMs can process and utilize for subsequent generation tasks.

\noindent\textbf{Multimodal Encoding~} Each text token is embedded into a vector \( e_{\text{{text}}} \in \mathbf{R}^d \), while the pretrained visual encoder transforms each input image into the feature \( e_{\text{{img}}} \in \mathbf{R}^{32 \times d} \). These embeddings are concatenated to create the input prompt features.

\noindent\textbf{Generative Vokens~} Since the original LLM’s \( V \) vocabulary only includes the textual tokens, we need to construct a bridge between the LLM and the generative model. Therefore, we introduce a set of special tokens \( V_{\text{{img}}} = \{[ \text{{IMG1}} ], [ \text{{IMG2}} ], \ldots, [ \text{{IMGn}} ]\} \) (by default $n=8$) as generative vokens into the LLM's vocabulary \( V \). The LLM's output hidden state for these vokens is harnessed for subsequent image generation, and the positions of these vokens can represent the insertion of the interleaved images. With all pretrained weights \( \theta_{\text{{pretrained}}} \) in MiniGPT-4 fixed, the trainable parameters include extra input embedding \( \theta_{\text{{voken\_input}}} \) and output embedding \( \theta_{\text{{voken\_output}}} \).

\noindent\textbf{Parameter-Efficient Fine-Tuning (PEFT)}
Parameter-efficient fine-tuning (PEFT)~\cite{houlsby2019parameter,hu2021lora,li2021prefix} is critical in training Large Language Models (LLMs), employed to adapt LLMs to downstream tasks without the need for extensive retraining. In PEFT, rather than updating all the parameters of a model, only a small subset of parameters is trained. This subset typically includes task-specific components or lightweight layers added to the original model architecture~\cite{tip_adapter,houlsby2019parameter,hu2021lora,dettmers2023qlora}. We apply PEFT to the MiniGPT-4~\cite{zhu2023minigpt} encoder, enhancing its ability to process and generate multimodal content based on given instructions or prompts. More specifically, this involves the use of prefix tuning~\cite{li2021prefix} and LoRA\cite{hu2021lora} over the entire language encoder -- \texttt{Vicuna}~\cite{vicuna2023} used in MiniGPT-4. Additionally, we implement learnable queries at the input of the transformer decoder, a conventional approach in sequence-to-sequence transformer architectures, to further improve the model's multimodal generation capabilities. We also adopted learnable queries at the input of the transformer decoder as a conventional setting for sequence-to-sequence transformer architectures~\cite{transformer}. Learnable queries in the decoder allow the model to have dynamic, adaptable representations for initiating the generation process. This is particularly useful when the model needs to generate outputs based on a mix of visual and textual inputs. Combined with the instruction tuning~\cite{training_lm}, it notably amplifies multimodal generation performance
across various datasets.

\subsection{Multimodal Generation Module}
To accurately align the generative vokens with the text-to-image generation models, we formulate a compact mapping module for dimension matching and incorporate several supervised losses, including causal language modeling loss and voken alignment loss. The causal language modeling loss assists the model in learning the correct positioning of tokens, while the voken alignment loss directly aligns the vokens with the appropriate conditional generation features of the diffusion model. Since the gradients of generative vokens' features can be directly calculated from images, shown on the right side of Fig.~\ref{fig:overview}, our method does not need comprehensive descriptions of images, leading to description-free learning.

\noindent\textbf{Voken Positioning~} 
We first jointly generate both text and vokens in the text space by following next-word prediction in autoregressive language model~\cite{vaswani2017attention}. During the training, we append the vokens $V_{\text{{img}}}$ to the positions of ground truth images and train the model to predict vokens within text generation. Specifically, the generated tokens are represented as \( W = \{w_1, w_2, \ldots, w_m\} \), where \( w_i \in V \cup V_{\text{{img}}} \), and the causal language modeling loss is defined as:
\begin{align}
L_{\text{{text}}}: & = - \sum_{i=1}^{m} \log  p(w_i|e_{\text{{text}}}, e_{\text{{img}}}, w_{1}, \ldots, w_{i-1}; \nonumber \\ & \theta_{\text{{pretrained}}}, \theta_{\text{{voken\_input}}}, \theta_{\text{{voken\_output}}}),\\ & \text{{ where }} w_i \in V \cup V_{\text{{img}}} \nonumber
\end{align}

\noindent\textbf{Voken Alignment for Image Generation~} 
Next, we align the output hidden state \( h_{\text{{voken}}} \), shown in Fig.~\ref{fig:overview}, with the conditional feature space of the text-to-image generation model. To map the voken feature \( h_{\text{{voken}}} \) to a feasible image generation conditional feature \( e_{\text{{text\_encoder}}} \in \mathbf{R}^{L \times \hat{d}} \) (where $L$ is the maximum input length of text-to-image generation text encoder, and $\hat{d}$ is the dimension of encoder output feature in text-to-image generation model). We construct a feature mapper module, including a two-layer MLP model \( \theta_{\text{{MLP}}} \), a four-layer encoder-decoder transformer model \( \theta_{\text{{enc-dec}}} \), and a learnable decoder feature sequence \( q \). The mapping feature \( \hat{h}_{\text{{voken}}} \) is then given by:
\begin{equation}
\hat{h}_{\text{{voken}}}: = \theta_{\text{{enc-dec}}}(\theta_{\text{{MLP}}}(h_{\text{{voken}}}), q) \in  \mathbf{R}^{L \times \hat{d}}
\end{equation}

To generate appropriate images, the mapping feature \( \hat{h}_{\text{{voken}}} \) is used as a conditional input in the denoising process. Intuitively, \( \hat{h}_{\text{{voken}}} \) should represent the corresponding conditional features that conduct the diffusion model to generate the ground truth image. We employ the latent diffusion model (LDM) loss as voken alignment loss for training the image generation module. During the training, the ground truth image is first converted to latent feature \( z_0 \) through the pretrained VAE (Variational Autoencoder)~\cite{vae}. Then, we obtain the noisy latent feature \( z_t \) by adding noise \( \epsilon \) to \( z_0 \). A pretrained U-Net model \( \epsilon_\theta \) is used to calculate the conditional LDM loss as:
\begin{equation}
L_{L D M}:=\mathbb{E}_{\epsilon \sim \mathcal{N}(0,1), t}\left[\left\|\epsilon-\epsilon_\theta\left(z_t, t, \hat{h}_{\text{{voken}}}\right)\right\|_2^2\right]
\end{equation}

To summarize, the causal language modeling loss enables the model to learn the accurate placement of tokens. Without this component, the model lacks the essential capability to predict when vokens should be generated during inference. Additionally, the voken alignment loss ensures the direct correspondence between vokens and the appropriate conditional generation characteristics of the diffusion model. In the absence of this loss, the model is unable to learn semantic vokens from images directly.
 This comprehensive approach ensures a coherent understanding and generation of both textual and visual elements, leveraging the capabilities of pretrained models, specialized tokens, and innovative training techniques.

\subsection{Training Strategy}
\label{sec:train}
Given the non-negligible domain shift between text and image domains, we observe that direct training on a limited interleaved text-and-image dataset can result in misaligning generated texts and images and diminished image quality.
Consequently, we adopt a two-stage training strategy: an initial pretraining stage focusing on coarse feature alignment for unimodal generation, followed by a fine-tuning stage dedicated to intricate feature learning for multimodal generation. Furthermore, to amplify the effectiveness of the generative tokens throughout the diffusion process, we incorporate the idea of classifier-free guidance~\cite{classifier_free_guidance} technique through the whole training process.

\noindent\textbf{Two-stage Training Strategy~} Recognizing the non-trivial domain shift between pure-text generation and text-image generation, we propose a two-stage training strategy: Pretraining Stage and Fine-tuning Stage. Initially, we align the voken feature with image generation features in single text-image pair datasets, such as CC3M, where each data sample only contains one text and one image, and the text is usually the caption of the image. During this stage, we utilize captions as LLM input, enabling LLM to generate vokens. Since these datasets include the image descriptive information, we also introduce an auxiliary loss to aid voken alignment, minimizing the distance between the generative feature \( \hat{h}_{\text{{voken}}} \) and the caption feature from the text encoder \( \tau_\theta \) in the text-to-image generation model:

\begin{equation}
L_{\text{{CAP}}} := \text{{MSE}}(\hat{h}_{\text{{voken}}}, \tau_\theta(c))
\end{equation}

The pretraining stage loss is expressed as \( L_{\text{{Pretrain}}} = \lambda_1 * L_{\text{{text}}} +  \lambda_2 * L_{\text{{LDM}}} +  \lambda_3 * L_{\text{{CAP}}} \), with selected values \( \lambda_1 = 0.01, \lambda_2 = 1, \lambda_3 = 0.1 \) to rescale the loss into a similar numerical range.

After the pretraining stage, the model is capable of generating images for single text descriptions but struggles with interleaved vision-and-language generation, which includes multiple text-image pairs and requires complicated reasoning for both text and image generation. To address this, in the fine-tuning stage, we further fine-tune our model with PEFT parameters by interleaved vision-and-language datasets, such as VIST, where the data sample has several steps with text-image and texts are sequentially relevant. During this stage, we construct three types of tasks from the dataset, encompassing (1) text-only generation: given the next image, generating the related text; (2) image-only generation: given the next text, generating the related image, and (3) multimodal generation: generating text-image pair by given context. The fine-tuning stage loss is given by \( L_{\text{{Fine-tune}}} = \lambda_1 * L_{\text{{text}}} +  \lambda_2 * L_{\text{{LDM}}} \). More implementation details can be found in Appendix~\ref{appendix:detail}.

\noindent\textbf{Classifier-Free Guidance (CFG)~} To enhance the coherence between the generated text and images, we first leverage the idea of Classifier-free Guidance for multimodal generation. Classifier-free guidance is introduced in the text-to-image diffusion process. This method observes that the generation model \( P_\theta \) can achieve improved conditional results by training on both conditional and unconditional generation with conditioning dropout. In our context, we want the model to focus directly on the output features \( h_{\text{{voken}}} \) from LLM. Instead of using original stable diffusion unconditional distributions (dropping \( \hat{h}_{\text{{voken}}} \)), the whole feature mapper also needs to be included during the unconditional process. Therefore, our objective is to accentuate the trainable condition \( h_{\text{{voken}}} \) and the generation model is fixed. During training, we replace \( h_{\text{{voken}}} \) with zero features \( h_0 \in \mathbf{0}^{n \times d} \) with a 10\% probability, obtaining the unconditional feature \( \hat{h}_0 = \theta_{\text{{enc-dec}}}(\theta_{\text{{MLP}}}(h_{0}), q) \). During inference, \( \hat{h}_0 \) serves as negative prompting, and the refined denoising process is:

\begin{equation}
\small
\begin{aligned}
& \log \widehat{\mathrm{P}_\theta}\left(\epsilon_t \mid z_{t+1}, \hat{h}_{\text{{voken}}}, \hat{h}_{0}\right) = \log \mathrm{P}_\theta\left(\epsilon_t \mid z_{t+1}, \hat{h}_{0}\right) +\\
& \gamma\left(\log \mathrm{P}_\theta\left(\epsilon_t \mid z_{t+1}, \hat{h}_{\text{{voken}}}\right) - \log \mathrm{P}_\theta\left(\epsilon_t \mid z_{t+1}, \hat{h}_{0}\right)\right)
\end{aligned}
\end{equation}

\section{Experiments}
To assess the efficacy of our model, we conducted a series of evaluations across multiple benchmarks. These experiments aim to address several key questions: (1) \emph{Can our model generate plausible images and reasonable texts?} (2) \emph{How does our model compare with state-of-the-art models in both single-turn and multi-turn interleaved vision-and-language generation tasks?} (3) \emph{What impact does the design of each module have on overall performance?} 
Below we will discuss the experimental setup and present a comprehensive analysis of our model's performance. 
We use three datasets: CC3M~\cite{sharma2018conceptual}, VIST~\cite{huang2016visual}, and MMDialog~\cite{feng2022mmdialog}. More details about datasets and data format can be found in Appendix~\ref{appendix:setting}.

\subsection{Implementation Details}
\label{appendix:detail}

In the pretraining stage, we introduce additional voken embeddings at both the input and output layers of the Vicuna-7B model, while keeping the embeddings of other tokens fixed. These new embeddings -- denoted as \( \theta_{\text{voken\_input}} \) and \( \theta_{\text{voken\_output}} \) -- along with the feature mapper module \( (\theta_{\text{MLP}}, \theta_{\text{enc\_dec}}, q) \) are jointly trained on the CC3M dataset, which consists of single text-image pairs. Training is conducted using the AdamW optimizer over two epochs, with a batch size of 48, amounting to over 110,000 steps, and a learning rate of \( 2 \times 10^{-4} \).

In the subsequent fine-tuning stage, we incorporate LoRA modules -- denoted as \( \theta_{\text{LoRA}} \) -- into Vicuna for the generation of both tokens and vokens. We keep the MLP model \( \theta_{\text{MLP}} \) and decoder query \( q \) fixed. The model is then fine-tuned on interleaved vision-and-language datasets, like VIST and MMDialog. The trainable parameters for this stage are \( \theta = \{\theta_{\text{voken\_input}}, \theta_{\text{voken\_output}}, \theta_{\text{LoRA}}, \theta_{\text{enc\_dec}} \} \). Training is carried out using the AdamW optimizer over four epochs, with a batch size of 32 and a learning rate of \( 2 \times 10^{-5} \). Trainable parameters are nearly 6.6 million, and all training can be completed on a server equipped with 4 A6000 GPUs.

\subsection{Experimental Setup}

\subsubsection{Baselines~} 
For a comprehensive evaluation of our performance in multimodal generation, we conducted comparative analyses with several prominent baseline models: the Fine-tuned Unimodal Generation Models, Two-stage Baseline, GILL~\footnote{Given the variations in the valid data within the CC3M dataset, we made adjustments to ensure fair comparisons. Specifically, we retrained it on our specific CC3M data, following the guidelines in their official implementation (\url{https://github.com/kohjingyu/gill}).}~\cite{koh2023generating}, and Divter~\citep{sun2021multimodal}:
\begin{itemize}
    \item \textbf{Fine-tuned Unimodal Generation Models}: To facilitate fair comparisons in both image and text generation, we fine-tuned two separate models, Stable Diffusion 2.1 and MiniGPT-4~\cite{zhu2023minigpt}, utilizing the VIST dataset. Within the Stable Diffusion 2.1~\cite{rombach2021highresolution} model, the U-Net parameters were fine-tuned. For MiniGPT-4's LLM part, LoRA parameters were fine-tuned.
    \item \textbf{Two-stage Baseline}: A common approach in multimodal generation involves first employing Large Language Models (LLMs) to create image captions, which are then fed into text-to-image models for image generation~\cite{wu2023visual}. We create such a two-stage baseline for comparison with our end-to-end method by fine-tuning MiniGPT-4 for caption generation and Stable Diffusion 2.1 for text-to-image generation. Given the absence of image descriptions in the VIST dataset, we incorporate a SOTA image captioning model, InstructBLIP-13B~\cite{instructblip}, to generate synthetic captions for supervision.
    \item \textbf{GILL}: GILL is a recent innovation that allows the LLM to generate vokens using a pre-trained text-to-image generation model for single-image generation, where GILL minimizes the Mean Squared Error (MSE) loss between the text-to-image text encoding feature and voken features, similar to \( L_{CAP}\) in our approach. For fine-tuning on multimodal datasets, since GILL requires image captions for training, we use Descriptions of Images-in-Isolation (DII)~\cite{huang2016visual} in the VIST fine-tuning and generate captions for MMDialog fine-tuning. Contrarily, \modelname does not related on all caption data during multimodal generation fine-tuning.
    \item \textbf{Divter}~\citep{sun2021multimodal}: Divter is a state-of-the-art conversational agent developed for multimodal dialogue contexts. It introduces a customized transformer structure for generating multimodal responses. Divter's methodology includes pretraining on a vast corpus of text-only dialogues and text-image pairs, followed by fine-tuning on a selected set of multimodal response data. The MMDialog dataset regards Divter's method as the baseline.
\end{itemize}

\begin{table}[t]
    \centering
    \begin{minipage}{.48\textwidth}
        \centering
        \begin{adjustbox}{width=0.95\textwidth,center}
            \small
            \begin{tabular}{lcc}
                \toprule
                Model & CLIP-I ($\uparrow$) & FID ($\downarrow$) \\
                \midrule
                SD 2.1~\cite{rombach2021highresolution} & 0.59 & 393.49 \\
                Fine-tuned SD 2.1 & 0.61 & 390.25 \\
                Two-stage Baseline & 0.57 & 403.06 \\
                GILL~\cite{koh2023generating} & 0.60 & 381.88 \\
                \modelname (Prefix Tuning) & 0.65 & 381.55 \\
                \modelname (LoRA) & \textbf{0.66} & \textbf{366.62} \\
                \bottomrule
            \end{tabular}
        \end{adjustbox}
        \caption[Image generation on VIST]{Image generation on VIST. Given the historical context, models need to generate images for each step. FID scores evaluate the visual diversities between generated and ground truth images within each story sequence.}
        \label{all_step_img}
    \end{minipage}
    \hfill
    \begin{minipage}{.48\textwidth}
        \centering
        \begin{adjustbox}{width=\textwidth,center}
            \small
            \begin{tabular}{lccc}
                \toprule
                Model & S-BERT ($\uparrow$) & Rouge-L ($\uparrow$) & Meteor ($\uparrow$) \\
                \midrule
                GILL~\cite{koh2023generating} & 0.3864 & 0.1784 & 0.1951 \\
                MiniGPT-4~\cite{zhu2023minigpt} & 0.6273 & \textbf{0.3401} & \textbf{0.3296} \\
                \modelname & \textbf{0.6315} & 0.3373 & 0.3263 \\
                \bottomrule
            \end{tabular}
        \end{adjustbox}
        \caption[Narration Generation on VIST]{Narration Generation on VIST. We added LoRA fine-tuning for GILL, MiniGPT-4, and \modelname with the same LoRA configuration. The results show that adding generative vokens does not hurt the performance on the multimodal comprehension tasks.}
        \label{all_step_lang}
    \end{minipage}
    \vspace{-2em}
\end{table}

\subsubsection{Metrics~}
To comprehensively assess the model performance across image, text, and multimodal dimensions, we employ a diverse set of metrics. For evaluating the quality and diversity of generated images, we utilize the Inception Score (IS)~\cite{salimans2016improved}, and Fréchet Inception Distance (FID)~\cite{heusel2017gans}. Textual performance is gauged through metrics such as BLEU~\cite{papineni2002bleu}, Rouge-L~\cite{lin2004rouge}, METEOR~\cite{banerjee2005meteor}, and Sentence-BERT (S-BERT) ~\cite{reimers2019sentence} scores.

From the multimodal perspective, we leverage CLIP-based metrics~\cite{rombach2021highresolution} to assess the similarities between generated content and ground truth. CLIP-I evaluates the similarity between generated and ground-truth image features. To address potential misalignments in the multimodal generation, such as when the ground truth is text-only, but the output is multimodal, we utilize MM-Relevance~\cite{feng2022mmdialog}. This metric calculates the F1 score based on CLIP similarities, providing a nuanced evaluation of multimodal coherence.

We also incorporate human evaluation to assess the model's performance. We examine the model's effectiveness from three perspectives: (1) \emph{Language Continuity}: assessing if the produced text aligns seamlessly with the provided context; (2) \emph{Image Quality}: evaluating the clarity and relevance of the generated image; and (3) \emph{Multimodal Coherence}: determining if the combined text-image output is consistent with the initial context.

\subsection{Main Results}

\begin{figure}[!t]
     \centering
     \includegraphics[width=0.48\textwidth]{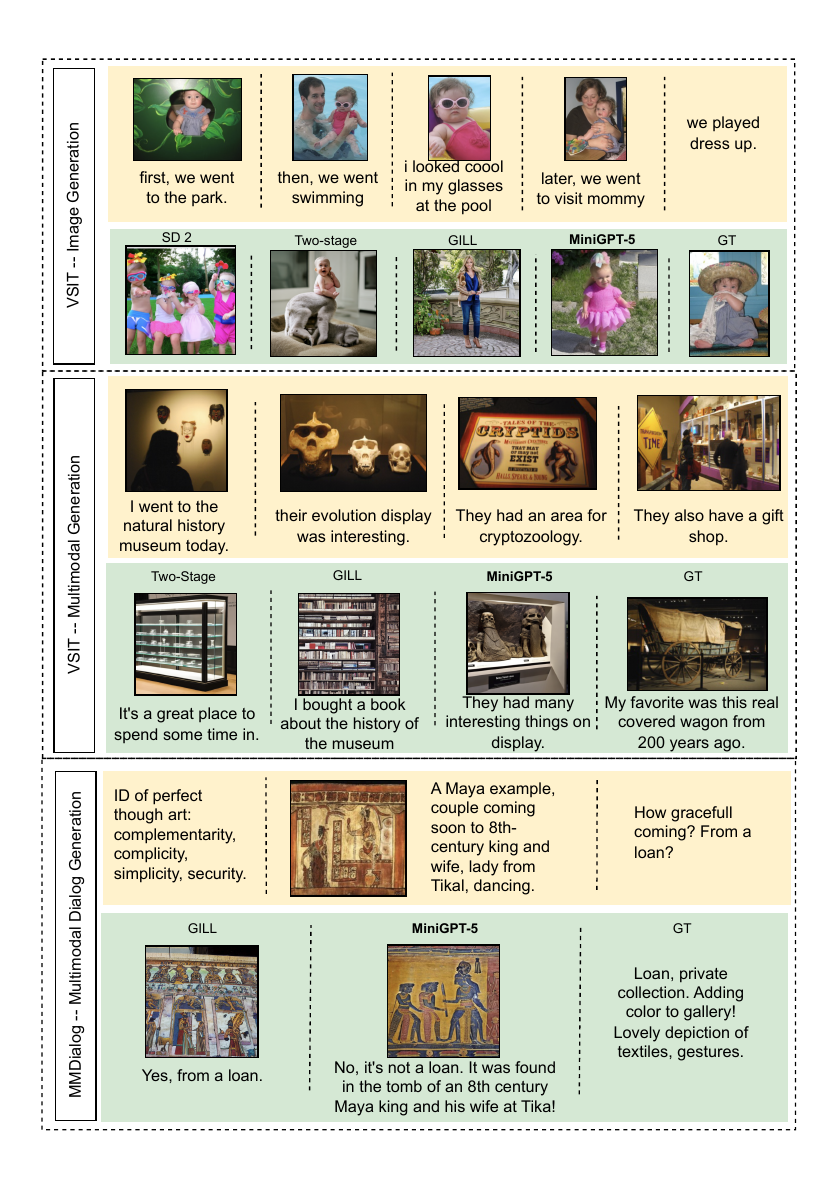}
     \caption{Qualitative examples from \modelname and baselines on the VIST and MMDialog datasets. The orange blocks indicate the input prompts, while the green blocks include model outputs. The comparisons show that \modelname can produce coherent and high-quality multimodal output. We would like to emphasize that \modelname does not use any caption data during fine-tuning on VIST and MMDialog, which obeys to our description-free settings. More qualitative examples can be found in the Appendix~\ref{appendix:more_qualitative}.}
     \label{fig:qualit}
     \vspace{-1em}
\end{figure}

In this subsection, we present the performance of different models on the VIST~\cite{huang2016visual} and MMDialg~\cite{feng2022mmdialog} datasets. Our evaluations span all vision, language, and multimodality domains to showcase the versatility and robustness of the proposed models.

\begin{table}[t]
    \centering
    \resizebox{\linewidth}{!}{
    \begin{tabular}{lccc}
        \toprule
        Model & \modelname & Two-stage Baseline & Tie \\
        \midrule
        Language Continuity (\%) & \textbf{55.22} & 34.89 & 9.89 \\
        Image Quality (\%) & \textbf{52.43} & 37.79 & 9.78 \\
        Multimodal Coherence (\%)& \textbf{56.90} & 28.88 & 14.22 \\
        \bottomrule
    \end{tabular}}
    \caption{VIST Human Evaluation on 5,000 samples for multimodal generation from Language Continuity, Image Quality, and Multimodal Coherence aspects. The results indicate, in more than 70\% cases, the \modelname is better or on par with the two-stage baseline.}
    \label{table:human_eval}
    \vspace{-15pt}
\end{table}

\vspace{1ex} \noindent\textbf{Unimodal Generation on VIST~}
To evaluate the model performance on image generation and text generation, we systematically provide models with prior history context and subsequently assess the generated images and narrations at each following step. Tables~\ref{all_step_img} and ~\ref{all_step_lang} outline the results of these experiments on the VIST validation set, showing the performance in both image and language metrics, respectively. The findings demonstrate that \modelname can generate coherent, high-quality images utilizing long-horizontal multimodal input prompts across all data, without compromising the original model's ability for multimodal comprehension, indicating the efficacy of our model in diverse settings.

\noindent\textbf{Multimodal Generation on VIST~}
To assess the quality of multimodal generation, we test both our model and the baselines on the VIST validation set by human evaluation. Given a preceding multimodal sequence, models are tasked with producing the subsequent scenario for each task. We select a random sample of 5,000 sequences, with each requiring evaluation by two workers. These evaluators are tasked with determining the superior multimodal output based on three criteria: Language Continuity, Image Quality, and Multimodal Coherence. This assessment is facilitated using Amazon Mechanical Turk~\cite{crowston2012amazon}, with a representative example (Fig.~\ref{fig:appendix_human}) provided in the Appendix. As depicted in Table~\ref{table:human_eval}, our model, \modelname, is found to generate more fitting text narrations in around 55\% of instances, deliver superior image quality in around 53\% of cases, and produce more coherent multimodal outputs in around 56\% of the scenarios. This data distinctly showcases its enhanced multimodal generation capabilities compared to the two-stage baseline, which must generate intermediate image captions first.


\noindent\textbf{Multimodal Dialog Generation on MMDialog~}
We conduct an evaluation of our method on the MMDialog dataset to determine the effectiveness of generating precise and appropriate multimodal information in multi-turn conversational scenarios. The model is required to generate either unimodal or multimodal responses based on the previous turns during the conversations. Our results, as presented in Table~\ref{table:mmdialog}, demonstrate that \modelname outperforms the baseline model Divter in terms of generating more accurate textual responses. While the image qualities of the generated responses are similar, \modelname excels in MM-Relevance compared to the baselines. This indicates that our model can better learn how to position image generation and produce highly coherent multimodal responses appropriately.

\begin{table}[t]
    \centering
    \resizebox{\linewidth}{!}{
    \begin{tabular}{lccccc}
          \toprule
        Model & IS ($\uparrow$) & BLEU-1 ($\uparrow$) & BLEU-2 ($\uparrow$) &Rouge-L ($\uparrow$) & MM-Relevance ($\uparrow$) \\
        \midrule
        Divter~\cite{sun2021multimodal} & 20.53 & 0.0944 & 0.0745 & 0.1119 & 0.62 \\
        GILL~\cite{koh2023generating} & \textbf{23.78} & 0.2912 & 0.1945 & \textbf{0.1207} & 0.64 \\
        \modelname  & 20.23 & \textbf{0.3369} & \textbf{0.2323} & 0.1176 & \textbf{0.67} \\
        \bottomrule
    \end{tabular}}
    \caption{Multimodal generation results on MMDialog test set. In order to compare with their baseline, we use the same metrics reported in MMDialog~\citep{feng2022mmdialog}.}
    \label{table:mmdialog}
\end{table}

\subsection{Ablation Studies}
 To further evaluate the effectiveness of our design, we conducted several ablation studies, and more ablation studies can be found in Appendix~\ref{appendix:cc3m_abalation}.
 
\begin{table}[t]
    \centering
    \resizebox{\linewidth}{!}{
    \begin{tabular}{lcccc}
        \toprule
        Model & CLIP-I ($\uparrow$) & CLIP-T ($\uparrow$) & IS ($\uparrow$) & FID ($\downarrow$) \\
        \midrule
        \modelname & \textbf{0.61} & \textbf{0.22} & \textbf{28.09} & \textbf{31.47} \\
        \modelname (w/o CFG) & 0.60 & 0.22 & 23.41 & 33.73 \\
        \modelname (w/o $L_{CAP}$) & 0.54 & 0.16 & 21.27 & 40.24 \\
        \modelname (w/o $L_{LDM}$) & 0.58 & 0.20 & 24.79 & 34.65 \\
        \bottomrule
    \end{tabular}}
    \caption{Evaluation of different method designs for image generation qualities on the CC3M validation set.}
    \label{table:ablation_cc3m}
    \vspace{-1em}
\end{table}

\noindent\textbf{Evaluation of Classifier-Free Guidance (CFG)~} To assess the effectiveness of the CFG strategy, we trained our model without CFG dropoff. During inference, the model utilized the original CFG denoising process, which utilized the empty caption feature from Stable Diffusion's text encoder as negative prompt features. The results in Table~\ref{table:ablation_cc3m} demonstrate that all metrics are worse without CFG, indicating that the CFG training strategy improves the image generation quality.

\noindent\textbf{Evaluation of Different Loss Guidance~} As described in Sec.~\ref{sec:train}, we introduced an auxiliary loss, denoted as $L_{CAP}$ for CC3M training. To assess the impact of this loss and determine if the single caption loss alone can generate high-quality images like GILL, we trained our model without the caption loss $L_{CAP}$ (alignment between the mapped generative voken features and the caption features from stable diffusion text encoder) and the conditional latent diffusion loss $L_{LDM}$ (alignment between the mapped generative voken features and conditional features for latent diffusion process of ground truth images) separately. The results, as shown in Table~\ref{table:ablation_cc3m}, indicate that the caption loss significantly aids in generating better images, and the voken alignment loss further enhances coherence and image quality performance.

\noindent\textbf{Influence of Input Types for Image Generation~}
To assess the impact of various types of input data for image generation, models are tasked with generating the final-step images based on specific prompts and comparing them with ground truth images by CLIP-I metric. All models are fine-tuned on data with full multimodal context and tested on various input types. As indicated in Table~\ref{single_step_vist}, the \modelname model exhibits exceptional proficiency in producing semantically precise images compared to other models. Furthermore, we observed increased CLIP similarities when more information was provided in the input, signifying the models' enhanced ability to process diverse, long-horizon multimodal inputs.

\begin{table}[t]
\resizebox{\linewidth}{!}{
    \centering
    \begin{tabular}{lcccc}
        \toprule
        Model & No Context & Text Context & Image Context & Image-Text Context \\
        \midrule
        SD 2~\citep{rombach2021highresolution} (Zero-shot) & 0.57 & 0.59 & - & -  \\
        GILL~\citep{koh2023generating} (Zero-shot) & 0.54 & 0.54 & 0.55 & 0.54  \\
        \modelname (Zero-shot) & 0.54 & 0.57 & 0.57 & 0.57  \\
        \hdashline
        Fine-tuned SD 2 & 0.59 & 0.61 & - & -  \\
        Two-stage Baseline & 0.54 & 0.56 & 0.57 & 0.58 \\
        \modelname (Prefix Tuning) & 0.60 & 0.63 & 0.68 & 0.70 \\
        \modelname (LoRA) & \textbf{0.61} & \textbf{0.64} & \textbf{0.69} & \textbf{0.70} \\
        \bottomrule
    \end{tabular}}
    \caption{Influence of prompts for image generation on CLIP-I metrics on VIST. We establish four distinct conditions for the final-step image generation: `No Context' (solely the last step's narration), `Text Context' (inclusive of historical textual narrations), `Image Context' (inclusive of historical images), and `Image-Text Context' (inclusive of both historical images and narrations). From the results, \modelname can generate more coherent images.} 
    \label{single_step_vist}
\end{table}

\noindent\textbf{Influence of Model Designs for Image Generation~}
To validate our feature mapper encoder/decoder architecture, we tested two alternatives: a model with \textbf{Fixed Queries}, where queries are initialized but not trained, and a \textbf{Decoder-Only} model, which pads its output for compatibility with the Stable Diffusion encoder.
The results of these experiments are detailed in Table~\ref{table:appendix_ablation_cc3m}. From the results of \modelname with fixed queries, we find there exists a slight trade-off between image-text coherence and image qualities, where fixed queries can lead to higher image metrics (IS and FID) but lower CLIP similarities. Meanwhile, \modelname consistently outperforms the Decoder-Only results in all four evaluation metrics, validating the robustness and efficacy of \modelname's transformer encoder/decoder architecture design.

\begin{table}[t]
    \centering
    \resizebox{\linewidth}{!}{
    \begin{tabular}{lcccc}
        \toprule
        Model & CLIP-I ($\uparrow$) & CLIP-T ($\uparrow$) & IS ($\uparrow$) & FID ($\downarrow$) \\
        \midrule
        \modelname & \textbf{0.61} & \textbf{0.22} & 28.09 & 31.47 \\
        \modelname (Fixed Queries) & 0.60 & 0.21 & \textbf{28.55} & \textbf{30.56} \\
         \modelname (Decoder-Only) & 0.58 & 0.20 & 24.74 & 34.88 \\
        \bottomrule
    \end{tabular}}
    \caption{Evaluation of different model designs for image generation qualities on the CC3M validation set.}
    \label{table:appendix_ablation_cc3m}
    \vspace{-1em}
\end{table}

\begin{table}[t]
    \centering
    \resizebox{\linewidth}{!}{
    \begin{tabular}{lcccc}
        \toprule
        & \multicolumn{2}{c}{\textbf{CC3M}} & \multicolumn{2}{c}{\textbf{VIST}} \\
        \cmidrule(lr){2-3} \cmidrule(lr){4-5}
        \textbf{Model} & \textbf{CLIP-I} ($\uparrow$) & \textbf{FID} ($\downarrow$) & \textbf{CLIP-I} ($\uparrow$) & \textbf{FID} ($\downarrow$) \\
        \midrule
        Stable Diffusion 2.1~\citep{rombach2021highresolution} & 0.64 & \textbf{26.39} & 0.59 & 393.49 \\
        Stable Diffusion 3~\cite{sd3} & \textbf{0.68} & 29.19 & 0.63 & 380.35 \\
        \hdashline
        \modelname~(MiniGPT-4 + SD~2.1) & 0.61 & 31.47 & 0.66 & 366.62 \\
        \modelname~(LLaVA-1.5 + SD~2.1) & 0.62 & 28.96 & 0.65 & 376.58 \\
        \modelname~(Qwen2.5-VL + SD3) & 0.64 & 29.31 & \textbf{0.70} & \textbf{340.20} \\
        \bottomrule
    \end{tabular}}
    \caption{Backbone/base-model comparison on CC3M and VIST.
    SD~2.1/SD3 are unimodal T2I baselines; \modelname rows are interleaved text–image generation.}
    \label{table:backbone_ablation}
    \vspace{-1em}
\end{table}

\noindent\textbf{Backbone and Base-Model Study}
We evaluate whether the interface of \modelname depends on a particular multimodal backbone or image generator.
We consider three MLLMs, MiniGPT-4~\cite{zhu2023minigpt}, LLaVA-1.5~\citep{liu2024improved}, and Qwen2.5-VL~\citep{qwen2_5_vl}, and two diffusion backbones, Stable Diffusion~2.1 (SD~2.1)~\citep{rombach2021highresolution} and Stable Diffusion~3 (SD3)~\cite{sd3}, shown in Table~\ref{table:backbone_ablation}.
On VIST, \modelname(Qwen2.5-VL+SD3) outperforms both SD3 alone and \modelname(MiniGPT-4+SD2.1), indicating that stronger backbones translate into better interleaved coherence through our voken-to-diffusion interface. With the generator fixed (SD~2.1), swapping MiniGPT-4 for LLaVA-1.5 yields comparable VIST results and a modest improvement on CC3M, suggesting robustness to the choice of MLLM. On CC3M, unimodal T2I baselines achieve the best single-turn extremes, while \modelname(Qwen2.5-VL+SD3) remains competitive and uniquely supports interleaved generation; the residual gap likely stems from the added mapping from LLM states to diffusion conditions and our objective emphasizing interleaving rather than single-turn T2I. Overall, gains on VIST persist across MLLMs and improve with a stronger generator, supporting that our contributions are orthogonal to specific backbones yet benefit from newer ones.

\section{Conclusion}
We introduce \modelname, designed to augment the capabilities of LLMs for multimodal generation by aligning the LLM with a pretrained text-to-image generation model. Our approach demonstrates substantial improvements. The limitation of \modelname is that we still find the object texture is hard to maintain in the new generation. Through this work, we aspire to set a new benchmark for existing and future multimodal generative models, opening doors to applications previously deemed challenging due to the disjointed nature of existing image and text synthesis paradigms.

{
    \small
    \bibliographystyle{ieeenat_fullname}
    \bibliography{main}
}

\clearpage
\appendix

\section{Experimental Settings}
\label{appendix:setting}
\subsection{Datasets} 
\vspace{1ex}\noindent\textbf{CC3M~\citep{sharma2018conceptual}:} Conceptual Captions (CC3M) dataset represents a remarkable collection of high-quality image captions, amassing approximately 3.3 million pairs of text and images from the internet. The CC3M dataset's diverse content, quality assurance, and support for multimodal learning make it a valuable asset for researchers and AI enthusiasts. Each dataset sample consists of an image accompanied by a corresponding text description, reflecting the richness of human language and visual perception. However, after accounting for license restrictions and eliminating invalid image links, the dataset comprises approximately 2.2 million data pairs suitable for training purposes and 10 thousand data pairs designated for validation.

\vspace{1ex}\noindent\textbf{VIST~\citep{huang2016visual}:} Visual Storytelling (VIST) dataset is an innovative compilation of visual narratives. The VIST dataset's engaging content, narrative structure, and emphasis on sequential understanding position it as an essential resource for researchers focusing on sequential image understanding. Each sequence within this dataset consists of five images accompanied by corresponding textual narratives, showcasing the intricate interplay between visual imagery and storytelling. Designed to foster creativity and challenge conventional image-captioning models, the dataset provides a platform for training and validating algorithms capable of generating coherent and contextually relevant stories. After eliminating the invalid image links, we got over 65 thousand unique photos organized into more than 34 thousand storytelling sequences for training and 4 thousand sequences with 8 thousand images for validation.

\vspace{1ex}\noindent\textbf{MMDialog~\citep{feng2022mmdialog}:} Multi-Modal Dialogue (MMDialog) dataset stands as the largest collection of multimodal conversation dialogues. The MMDialog dataset's extensive scale, real human-human chat content, and emphasis on multimodal open-domain conversations position it as an unparalleled asset for researchers and practitioners in artificial intelligence. Each dialogue within this dataset typically includes 2.59 images, integrated anywhere within the conversation, showcasing the complex interplay between text and visual elements. Designed to mirror real-world conversational dynamics, the dataset is a robust platform for developing, training, and validating algorithms capable of understanding and generating coherent dialogues that seamlessly blend textual and visual information.

\subsection{Data Format} 
\vspace{1ex}\noindent\textbf{Pretraining Stage} In the pretraining stage, we aim to synchronize the generative voken with the text-to-image model's conditional feature, focusing on single-turn text-image pairs. To achieve this, we utilize data from the CC3M dataset, constructing training samples by appending vokens as image placeholders after the captions, such as “a big black dog [IMG1] … [IMGn].” The Language Model (LLM) is then tasked with only generating these placeholders for text creation, and the corresponding output hidden features are further employed to compute the conditional generation loss with the ground truth image.

\vspace{1ex}\noindent\textbf{Fine-tuning Stage} In this stage, we utilize the VIST and MMDialog datasets, which contain multi-turn multimodal data. During training, we integrate placeholders for input images, such as '$<$Img$>$$<$ImageHere$>$$<$/Img$>$', into the input text prompts when applicable. These prompts also encompass various instructions corresponding to different task types, with outputs manifesting as pure-text, pure-voken, or text-voken combinations. Below, we present example templates in the VIST dataset to illustrate the different task types:
\begin{itemize}
    \item \textbf{Text Generation:} Input: “$<$History Context$>$ What happens in the next scene image: $<$Img$>$$<$ImageHere$>$$<$/Img$>$”; Output: “$<$Text Description$>$”
    \item \textbf{Image Generation:} Input: “$<$History Context$>$ Generate an image with the scene description: [Text Description]”; Output: “[IMG1]...[IMGn]”
    \item \textbf{Text-Image Generation:} Input: “$<$History Context$>$ What should happen then?”; Output: “$<$Text Description$>$ [IMG1]...[IMGn]”
\end{itemize}
By structuring the input and output in this manner, we create a flexible framework that accommodates various multimodal tasks, enhancing the model's ability to interpret and generate textual and visual content. The history context in the VIST dataset includes all previous story steps with texts and images. In the MMDialog dataset, due to the limitation of computational resources, we only use up to one previous turn as the history context, and all data are formatted into the dialog.

\begin{figure*}[!t]
     \centering
     \includegraphics[width=0.8\textwidth]{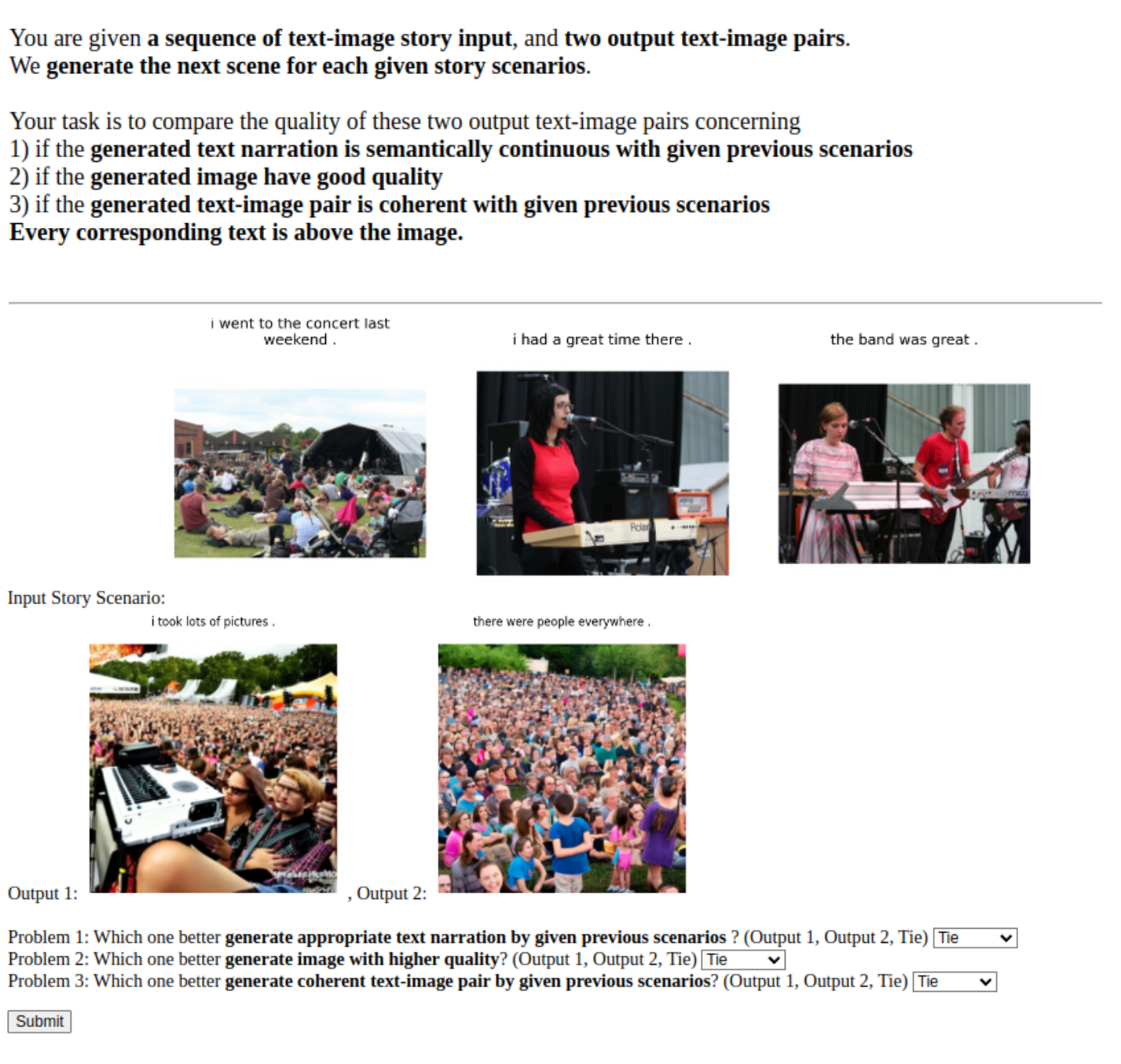}
     \caption{Screenshot for human evaluation interface on the Amazon Mechanical Turk crowdsource evaluation platform. Output 1 is generated by \modelname, while output 2 is generated by the two-stage baseline. }
     \label{fig:appendix_human}
\end{figure*}

\section{More Experiments}
\label{appendix:cc3m_abalation}
\subsection{Evaluation of Guidance Scale} 
\begin{figure*}[t]
  \centering
  \subfloat[FID vs CFG Scale]{\includegraphics[width=0.4\textwidth]{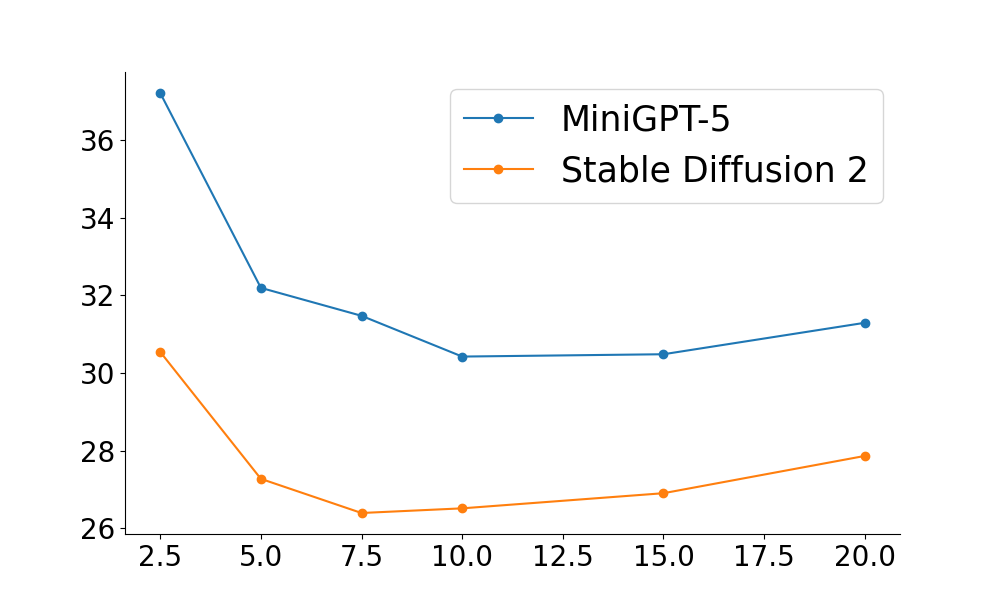}}
  \subfloat[IS vs CFG Scale]{\includegraphics[width=0.4\textwidth]{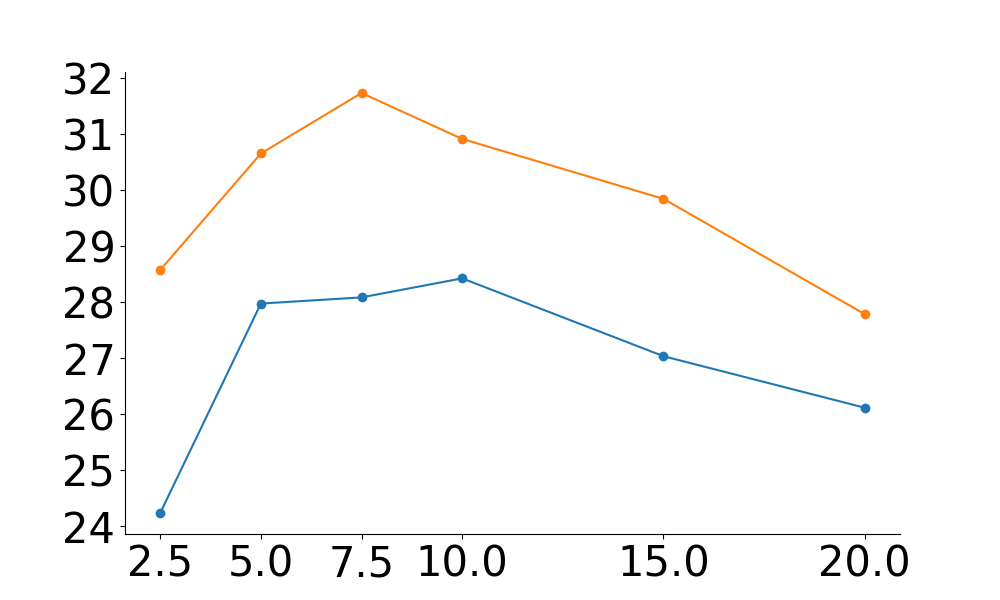}}\\
  \subfloat[CLIP-T vs CFG Scale]{\includegraphics[width=0.4\textwidth]{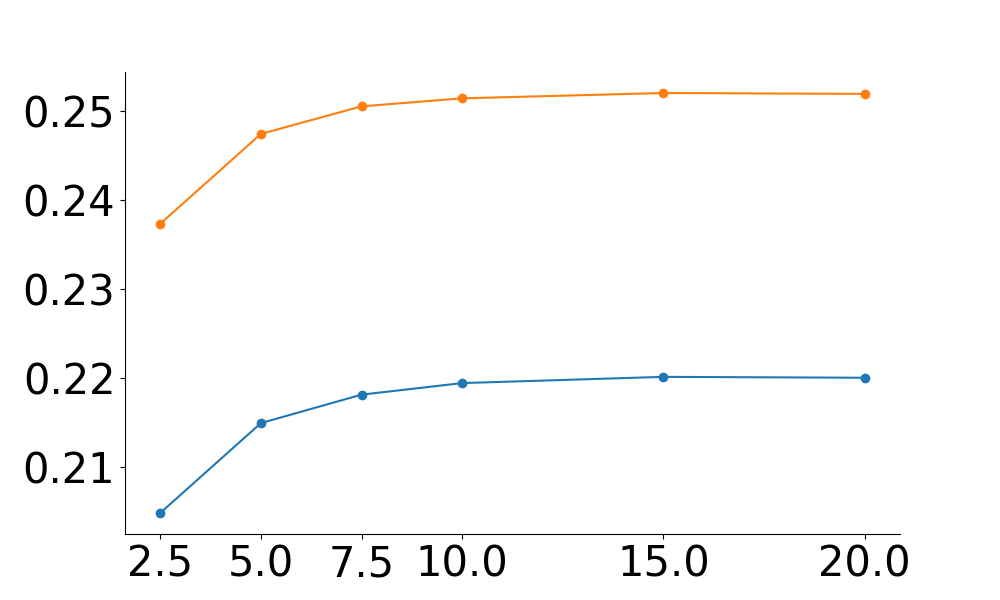}}
  \subfloat[CLIP-I vs CFG Scale]{\includegraphics[width=0.4\textwidth]{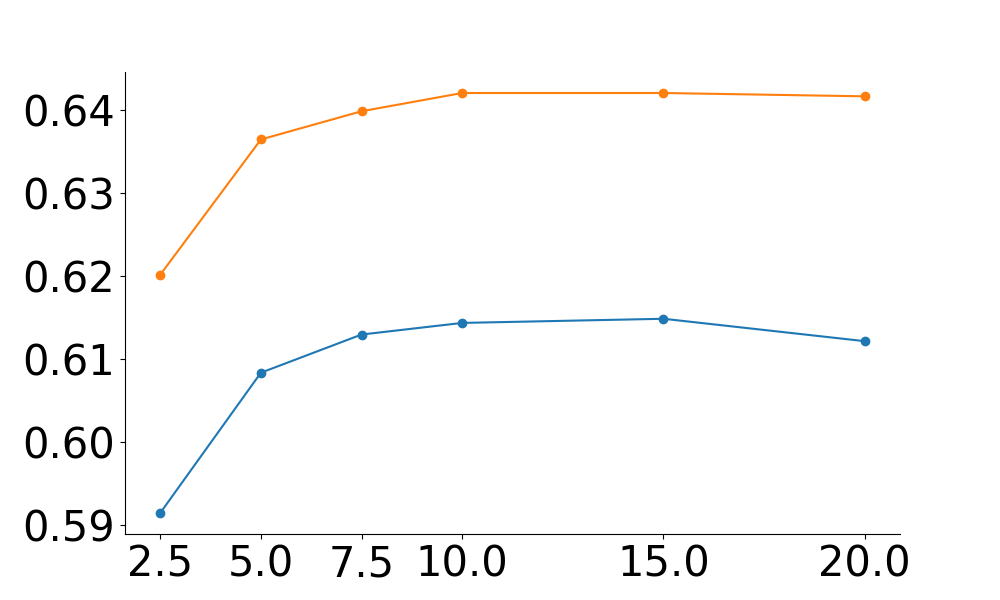}}
  \caption{Line charts for various metrics vs Classifier-free Guidance (CFG) scale on CC3M. The results suggest that our CFG strategy can exhibit comparable effectiveness to the CFG strategy employed in SD2, with the appropriate CFG scale significantly enhancing both image quality and coherence. }
  \label{fig:cc3m_scale}
\end{figure*}

Since our model incorporates CFG, evaluating how different guidance scales affect image generation is crucial. Therefore, we plotted several line charts in Fig~\ref{fig:cc3m_scale} to depict the changes in metrics with varying guidance scales. The figures reveal that the stable diffusion model and our model generate better images as the guidance scale increases. However, when the scale exceeds 10, the image semantic coherence stabilizes while the image quality declines. This suggests that the guidance scale should be set within a reasonable range for optimal image generation.

\subsection{Evaluation of Voken Number} 
\begin{figure*}[t]
  \centering
  \subfloat[FID vs $n_{voken}$]{\includegraphics[width=0.4\textwidth]{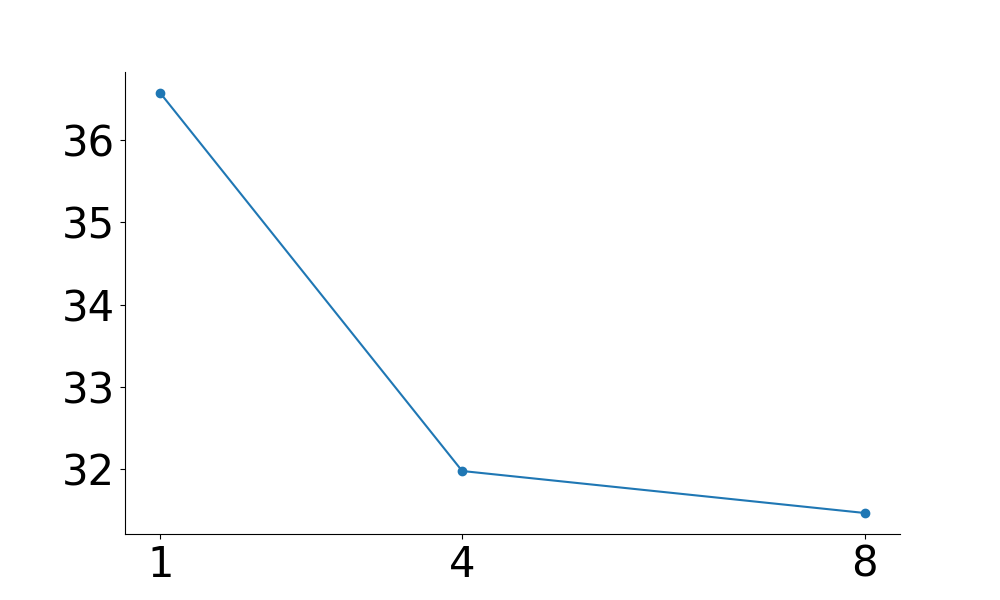}}
  \subfloat[IS vs $n_{voken}$]{\includegraphics[width=0.4\textwidth]{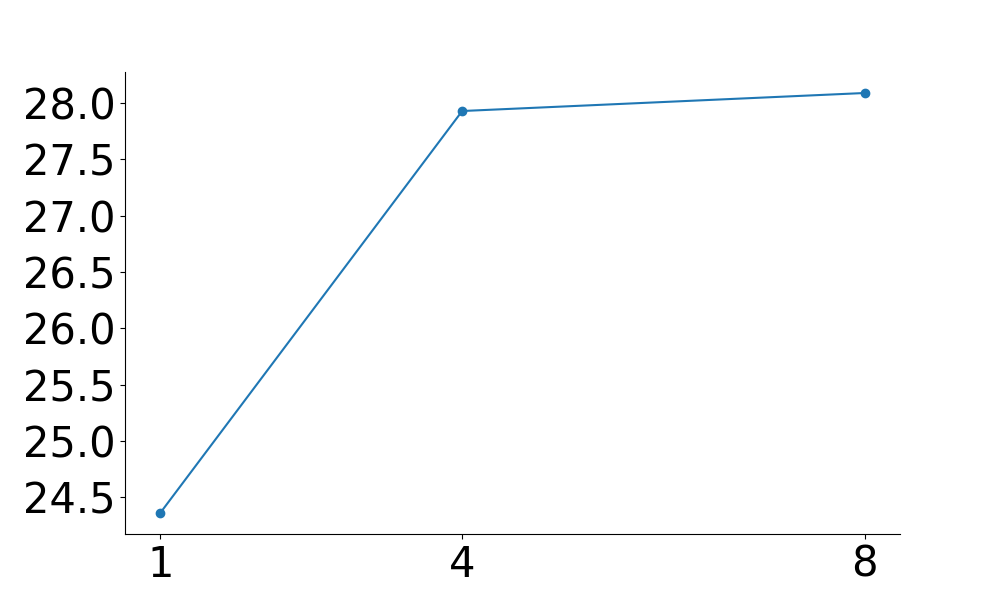}} \\
  \subfloat[CLIP-T vs $n_{voken}$]{\includegraphics[width=0.4\textwidth]{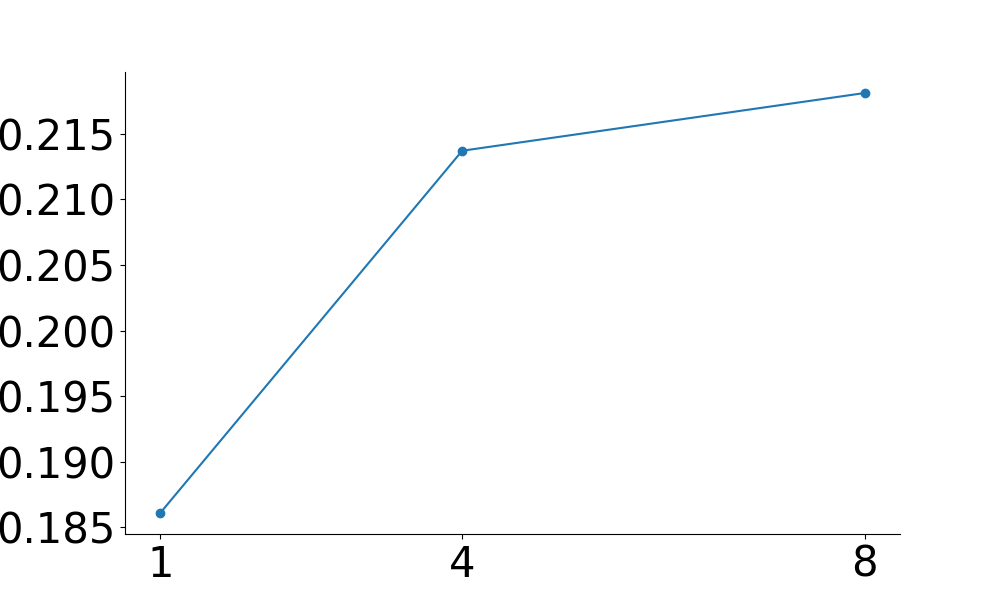}}
  \subfloat[CLIP-I vs $n_{voken}$]{\includegraphics[width=0.4\textwidth]{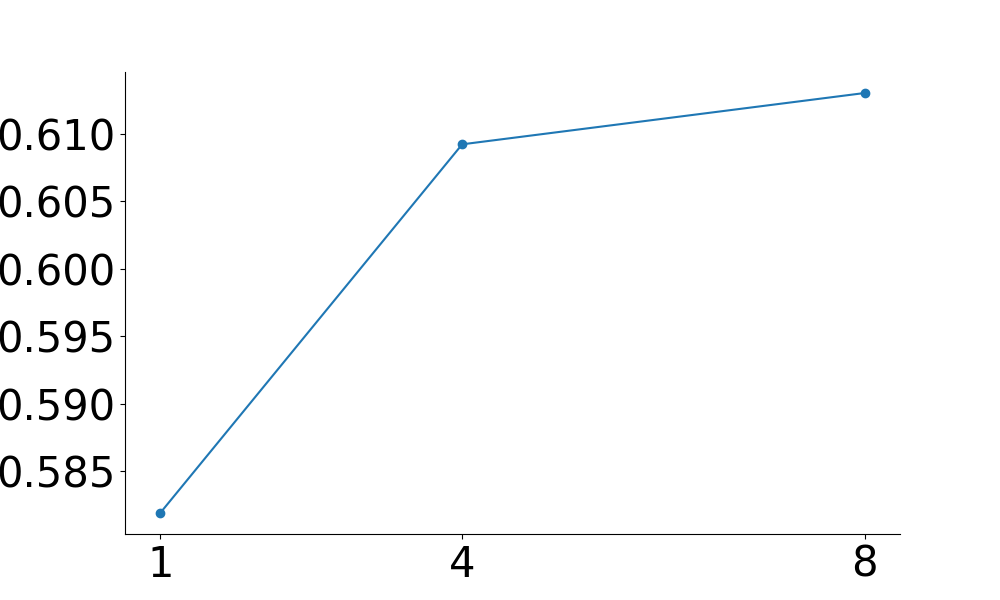}}
  \caption{Line charts for various metrics vs the number of vokens on CC3M. As the number of vokens increases, the image quality and CLIP scores improve. In this work, our default voken number is 8.}
  \label{fig:cc3m_num}
\end{figure*}

The voken features in our model are directly utilized as conditions in the text-to-image model, leading to the expectation that an increase in the number of vokens would enhance the model's representative capabilities. To validate this hypothesis, we experimented by training the model with varying numbers of vokens, ranging from 1 to 8. As illustrated in Fig~\ref{fig:cc3m_num}, the model's performance consistently improves with adding more vokens. This improvement is particularly noticeable when the number of vokens is increased from 1 to 4, highlighting the significant role that vokens play in enhancing the model's effectiveness.

\subsection{Texture Preservation via FID}
To assess whether models preserve the textures present in prior images when generating the last image of a story, we evaluate on VIST in a “final–step” setting: for each story, all preceding step images and narrations are provided as context, and the model must produce the final image. We then compute FID between the set of generated final images and the ground-truth final images from VIST. This directly measures the realism and low-level appearance consistency of the predicted finals relative to the true finals under an identical multimodal context, shown in Table~\ref{tab:fid_texture_vist}. In this setting, ViLGen (Qwen2.5-VL + SD3) attains the lowest FID, indicating stronger retention of low-level textures in the generated final images.

\begin{table}[t]
  \centering
  \resizebox{0.8\linewidth}{!}{
  \begin{tabular}{lc}
    \toprule
    \textbf{Model} & \textbf{FID ($\downarrow$)} \\
    \midrule
    GILL~\cite{koh2023generating} & 61.85 \\
    \modelname~(MiniGPT-4 + SD 2.1) & 59.48 \\
    \modelname~(Qwen2.5-VL + SD3) & \textbf{56.32} \\
    \bottomrule
  \end{tabular}}
  \caption{Texture preservation on VIST (final–step).}
  \label{tab:fid_texture_vist}
\end{table}
\vspace{-1em}

\section{More Qualitative Examples}
\label{appendix:more_qualitative}
\begin{figure*}[!t]
     \centering
     \includegraphics[width=0.9\textwidth]{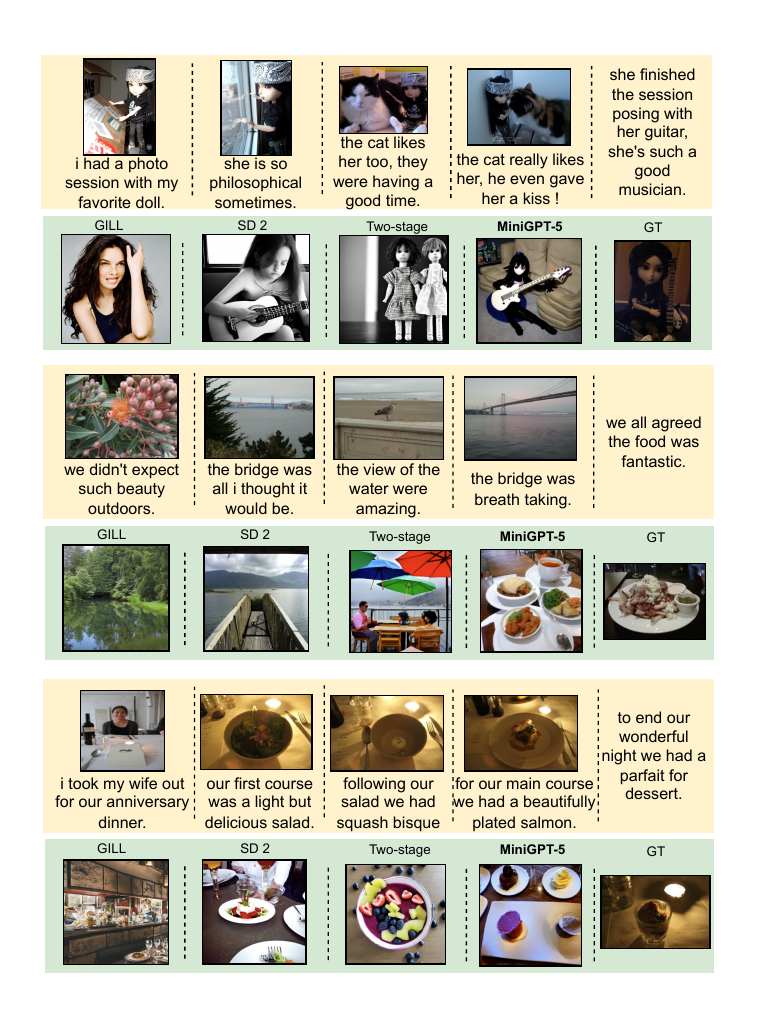}
     \caption{Comparative examples from \modelname and baselines on the VIST validation set for image generation with multimodal input. Orange blocks denote input prompts, while green blocks show model outputs.}
     \label{fig:appendix_vist}
\end{figure*}

\begin{figure*}[!t]
     \centering
     \includegraphics[width=0.75\textwidth]{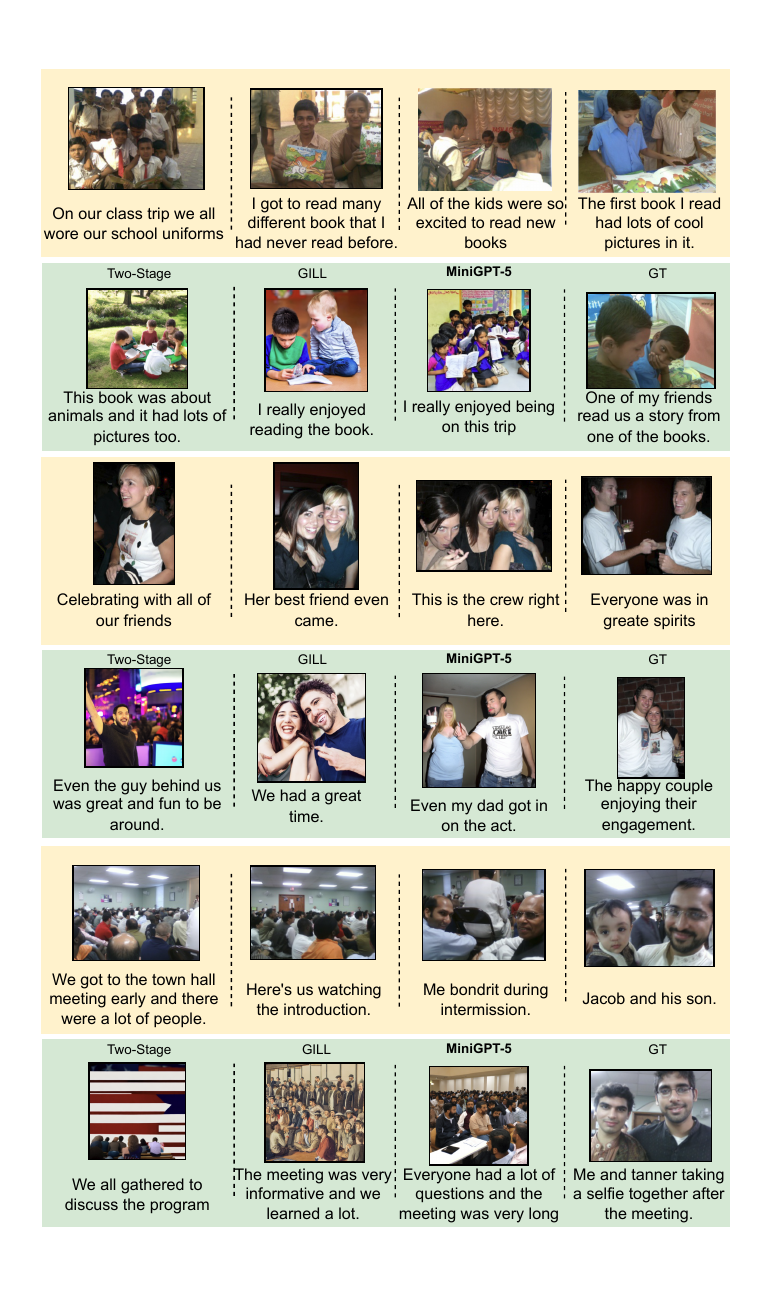}
     \caption{More qualitative examples from \modelname and baselines on VIST validation set for free multimodal generation. }
     \label{fig:appendix_vist2}
\end{figure*}

\begin{figure*}[!t]
     \centering
     \includegraphics[width=0.85\textwidth]{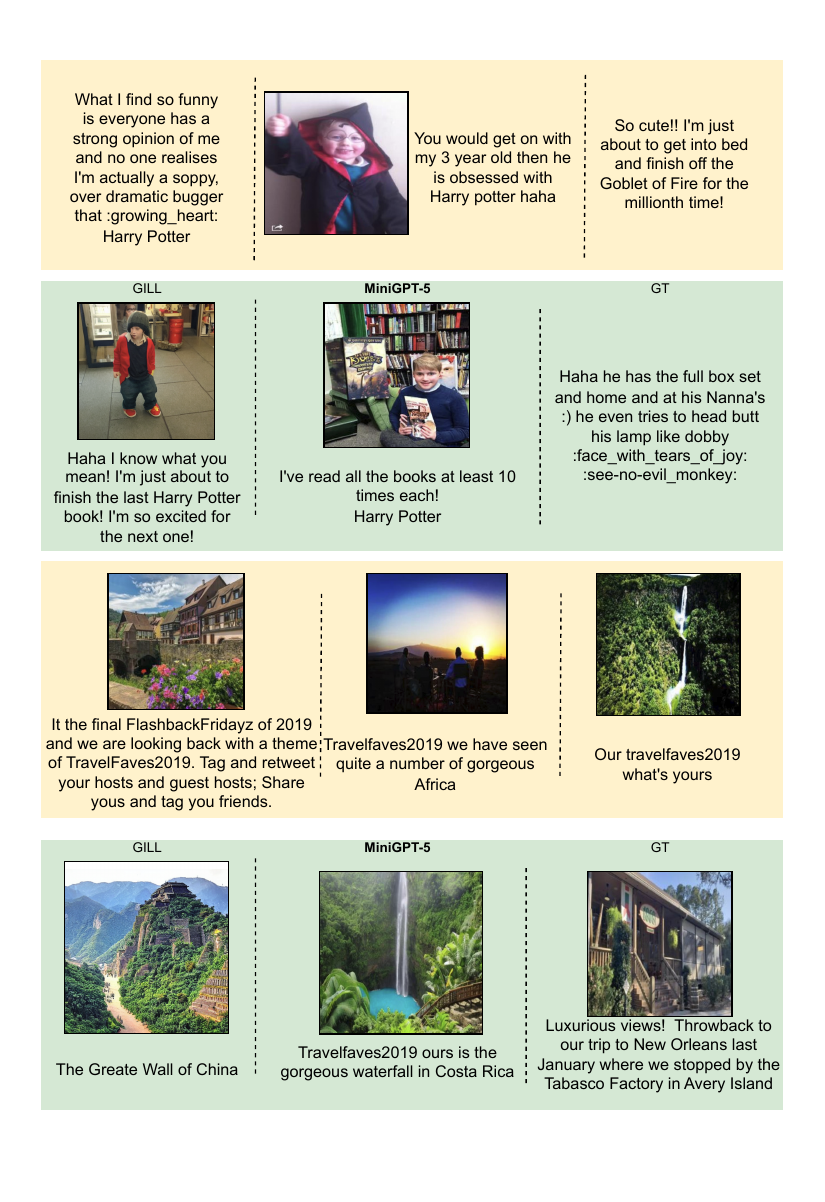}
     \vspace{-2em}
     \caption{More qualitative examples from \modelname on MMDialog test set for free multimodal dialog generation. }
     \label{fig:appendix_mmdialog}
\end{figure*}

\begin{figure*}[!t]
     \centering
     \includegraphics[width=\textwidth]{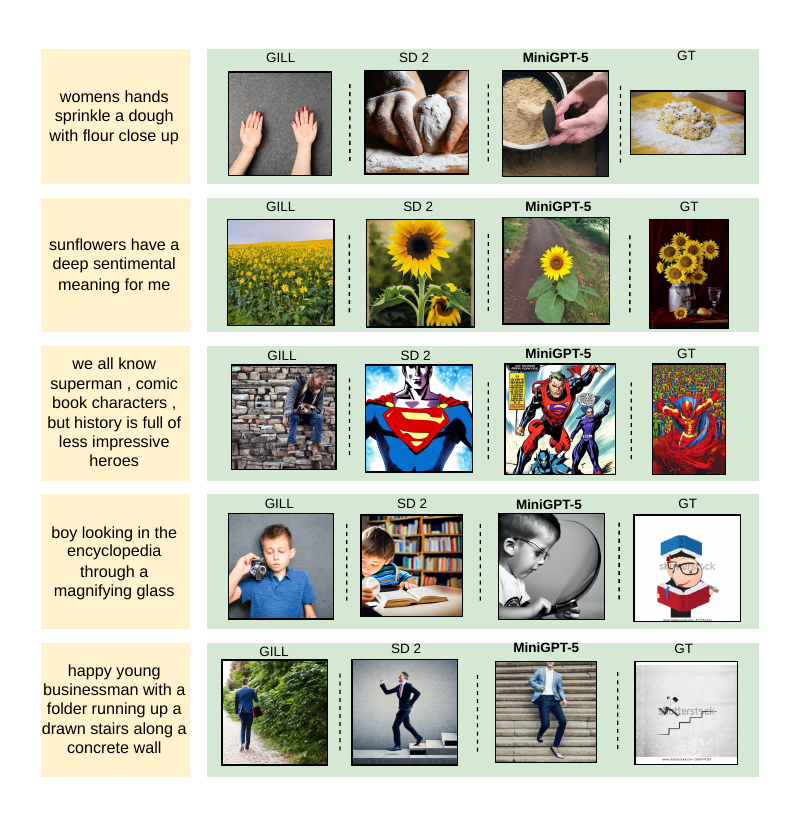}
     \caption{More qualitative examples from \modelname and baselines on CC3M validation set for single text-to-image generation. }
     \label{fig:appendix_cc3m}
\end{figure*}

In this section, we provide additional qualitative examples to further demonstrate the capabilities of \modelname. Figures~\ref{fig:appendix_vist},\ref{fig:appendix_vist2},\ref{fig:appendix_mmdialog}, and~\ref{fig:appendix_cc3m} showcase these examples across various datasets and settings.

Figure~\ref{fig:appendix_vist} presents a comparative analysis on the VIST validation set, illustrating how \modelname outperforms baseline models in terms of image generation quality and alignment with multimodal inputs. The examples highlight the superiority of \modelname in generating images that closely match the given text prompts.

In Figure~\ref{fig:appendix_vist2}, we focus on the performance of \modelname in free multimodal generation scenarios. The results clearly indicate an improvement over the Two-Stage baseline, emphasizing \modelname's ability to perform consistent and creative multimodal generation.

Figure~\ref{fig:appendix_mmdialog} showcases the application of \modelname in the context of the MMDialog test set. Here, the emphasis is on free multimodal dialog generation, with \modelname displaying a decent performance in generating coherent and contextually relevant multimodal dialogues.

Lastly, Figure~\ref{fig:appendix_cc3m} highlights \modelname's performance in single text-to-image generation tasks on the CC3M validation set. The examples underline the model's proficiency in generating visually accurate and contextually appropriate images from textual descriptions, surpassing the performance of baseline models.

Each figure includes a clear depiction of input prompts (indicated in orange blocks) and the corresponding model outputs (in green blocks), providing a comprehensive view of \modelname's capabilities across different multimodal generation tasks.

\end{document}